\documentclass[runningheads]{llncs}
\usepackage{amsmath}
\usepackage{multirow}
\usepackage{booktabs}
\usepackage{bm}
\newcommand{\rowheight}{\rule{0pt}{10pt}}
\usepackage{xcolor}
\usepackage{siunitx}
\usepackage{graphicx}
\usepackage{subcaption}
\usepackage{hyperref}

\begin{document}
\title{Exploring Federated Learning for Thermal Urban Feature Segmentation -- A Comparison of Centralized and Decentralized Approaches}
\titlerunning{ }
\author{
     Leonhard Duda\inst{1}\orcidID{0009-0008-2432-9392}\thanks{Code available in: \url{https://github.com/ai4os-hub/thermal-urban-feature-segmenter.git}}  \and
    Khadijeh Alibabaei \inst{1}\orcidID{0000-0002-2319-8211} \and
    Elena Vollmer\inst{2}\orcidID{0000-0002-8805-3726}\and
    Leon Klug\inst{2}\and
    Valentin Kozlov\inst{1}\orcidID{0000-0002-8770-3619}\and
    Lisana Berberi \inst{1}\orcidID{0000-0002-7632-2466}\and
    Mishal Benz\inst{1}\orcidID{0009-0006-1491-7567}\and 
    Rebekka  Volk  \inst{2}\orcidID{0000-0001-9930-5354}\and
    Juan Pedro Guti{\'e}rrez Hermosillo Muriedas  \inst{1}\orcidID{0000-0001-8439-7145}\and
    Markus Götz  \inst{1}\orcidID{0000-0002-2233-1041}\and
    Judith Sáinz-Pardo Díaz  \inst{3}\orcidID{0000-0002-8387-578X} \and
    Álvaro López García \inst{3}\orcidID{0000-0002-0013-4602} \and
    Frank Schultmann   \inst{2}\orcidID{0000-0001-6405-9763}\and
    Achim Streit  \inst{1}\orcidID{0000-0002-5065-469X}
}
\authorrunning{L. Duda et al.}
\institute{Scientific Computing Center (SCC), Karlsruhe Institute of Technology (KIT), Eggenstein-Leopoldshafen, Germany\\
\email{leonhard.duda}@kit.edu\\
%\url{http://www.springer.com/gp/computer-science/lncs} 
\and
Institute for Industrial Production
 (IIP), Karlsruhe Institute of Technology (KIT), Karlsruhe, Germany \\
\and 
Instituto de Física de Cantabria (IFCA), CSIC-UC, Avda. los Castros s/n, Santander, Spain
}
\maketitle              % typeset the header of the contribution
\begin{abstract}
Federated Learning (FL) is an approach for training a shared Machine Learning (ML) model with distributed training data and multiple participants. FL allows bypassing limitations of the traditional Centralized Machine Learning (\texttt{CL}) if data cannot be shared or stored centrally due to privacy or technical restrictions -- the participants train the model locally with their training data and do not need to share it among the other participants. This paper investigates the practical implementation and effectiveness of FL in a real-world scenario, specifically focusing on unmanned aerial vehicle (UAV)-based thermal images for common thermal feature detection in urban environments. The distributed nature of the data arises naturally and makes it suitable for FL applications, as images captured in two German cities are available. This application presents unique challenges due to non-identical distribution and feature characteristics of data captured at both locations. The study makes several key contributions by evaluating FL algorithms in real deployment scenarios rather than simulation. We compare several FL approaches with a centralized learning baseline across key performance metrics such as model accuracy, training time, communication overhead, and energy usage. This paper also explores various FL workflows, comparing client-controlled workflows and server-controlled workflows. %To this end, we use the perun package to measure and report the energy usage of each method. 
The findings of this work serve as a valuable reference for understanding the practical application and limitations of the FL methods in segmentation tasks in UAV-based imaging.

\end{abstract}
\section{Introduction}
Deep Learning (DL) has revolutionized many fields in science and industry, enabling significant advances in tasks like object detection and segmentation~\cite{10098596,9356353}. 
DL models, fueled by large datasets, have made it possible to develop systems that perform tasks with human-like accuracy. They are now applied across a variety of industries including healthcare, city management, and autonomous driving ~\cite{Awasthi2024.02.28.24303482,https://doi.org/10.1002/rob.21918,wu2024deep}.

Traditionally, these models are trained with a large amount of data stored in a central location. However, in response to growing data protection laws such as the General Data Protection Regulation (GDPR) and the EU AI act~\cite{GDPR2016}, the question of whether the traditional approach of storing and processing all data centrally is sustainable arises.
On the other hand, if the training data is distributed across multiple locations and sharing is hindered due to privacy or resource constraints, successful training becomes challenging, as there may not be enough data available at some of these locations, leading to poor model performance or failure to generalize over unseen data. 

To achieve the same predictive performance when the data is spread across multiple locations, Federated Learning (FL) was introduced in 2017~\cite{pmlr-v54-mcmahan17a}.
With this approach, the training data does not have to be shared, but a Machine Learning (ML) model can still be trained collaboratively and in a distributed manner. In the initial introduction of FL, there is a central server that aggregates updates of models trained on decentralized devices. Each device trains a model using its local data and sends only the model updates (not the raw data) to the central server. The server then combines local model weights to update the global model weights. This process enables collaborative model training while reducing the need to share raw data; however, information leakage risks may still exist without additional privacy safeguards such as differential privacy~\cite{Chang2018-iv,Warnat-Herresthal2021}. %This preserves privacy while taking advantage of distributed data sources. When the server is not trusted by the clients, decentralized FL was introduced to eliminate the need for sending model weights to the server. In this case, the communication occurs in a peer-to-peer manner and the tasks of the server are shifted to the clients \cite{Chang2018-iv,Warnat-Herresthal2021}.

The main objective of this paper is to investigate the effectiveness of FL in real-world applications and scenarios, with a particular focus on a segmentation task in unmanned aerial vehicle (UAV)-based imaging. Although FL has shown promising results in research papers, primarily in simulation scenarios, its practical performance in real-world settings should be explored, considering the limitations and challenges associated with real-world implementation~\cite{Agripina2024}.  This study aims to bridge this gap by evaluating the effectiveness of FL in a unique real-world use case: the detection of thermal anomalies in urban environments using UAV-based imaging. This application, part of the AI4EOSC project~\cite{ai4eosc}, involves detecting thermal anomalies caused by false alarms from urban features which complicates the detection of locations requiring maintenance or improvement. By reducing false positives, the system supports more efficient optimization of energy-related systems~\cite{MayerEpperleinVollmer2023_1000154253,VOLLMER2024105709,VollmerVolkSchultmann2023_1000163771}. 

A key feature of this work is the use of a real-world dataset collected from two cities, Munich (MU) and Karlsruhe (KA), in Germany. This dataset is characterized by significant imbalances and non-IID (non-independent and identically distributed) data across locations, posing unique challenges for both training and aggregation methods in FL. 

The main contributions of this paper are as follows: \begin{itemize} 
\item Highlight the real-world challenges associated with training FL models on non-IID, imbalanced data from multiple locations using UAV devices, showcasing the uniqueness of the data and its impact on FL performance.
\item Introduce and evaluate  FL aggregation strategies tailored to handle non-IID data in segmentation tasks on the UAV images.
\item Compare the performance of FL and Centralized Learning (\texttt{CL}) on a segmentation task in a real-world UAV imaging application, emphasizing practical challenges such as model accuracy, training time, and memory consumption. 
\item Investigate the performance of decentralized FL as a potential solution for privacy-preserving collaboration, comparing it against centralized FL in a real-world setting. 
\item Measure energy consumption and convergence time for both FL and \texttt{CL}, addressing the resource-constrained devices used in the real-world application and evaluating their feasibility in a practical deployment. 
\item  Explore the challenges and opportunities of using HPC systems as clients in FL, considering their potential to enhance computational capabilities.
\end{itemize}

As most of the experiments are done in real-world scenarios, this paper can be used as a valuable reference for understanding the practical applications and performance of FL in real-world settings.

The paper is organized as follows: Section~\ref{sec1} reviews the related work in the research areas of FL algorithms and workflows. Section~\ref{sec2} covers the methodology and setup, explaining our concrete implementation of a FL workflow based on the use case described, including a comparison and evaluation of all workflows with each other and with the centralized approach. The obtained results of applying different approaches are evaluated and discussed in Section~\ref{sec3}.

\section{Related Work}\label{sec1}

FL has gained significant traction in domains such as medical imaging and autonomous driving due to stringent privacy requirements~\cite{Banabilah:2022}.
In traditional distributed learning, it is assumed that all data comes from the same distribution, ensuring consistency across all participating nodes. But in real-world FL, data is typically non-IID, meaning that data across different clients can have different distributions, which poses additional challenges for model training and performance~\cite{9464278}.  While the FL approach is effective in simulation environments, it poses challenges in real-world applications due to this non-IID nature of the data and privacy concerns~\cite{Zili:2024,Wang:2024}.  To solve these issues, researchers have proposed improvements to aggregation algorithms in FL such as  \texttt{FedProx}~\cite{li2020federatedoptimizationheterogeneousnetworks}, Federated Optimization (\texttt{FedOpt})~\cite{reddi2021adaptive}, \texttt{Scaffold} \cite{karimireddy2021scaffoldstochasticcontrolledaveraging}, and \texttt{FedBN}~\cite{li2021fedbn}. 

FL  has been successfully used mostly in domains like medical imaging and autonomous driving primarily due to concerns about data privacy in these fields~\cite{10316635,GUAN2024110424}. In autonomous driving, FL is used to train a semantic segmentation model on data from cars in different cities, achieving robust performance \cite{10203444}. In medical image segmentation, FL has enabled institutions to build combined models that generalize better across patient populations. For example, \cite{MANTHE2024103270} used a FL model for brain tumor segmentation utilizing data from multiple hospitals enhancing the performance and generalizability of the segmentation. Furthermore,   \cite{Rashidi2024} evaluates several FL aggregation strategies, including \texttt{FedProx}, \texttt{FedMOON} ~\cite{9578660}, and \texttt{FedDC} ~\cite{kamp2023federatedlearningsmalldatasets}, alongside \texttt{FedAvg}, under both IID and non-IID scenarios in  Medical Object Detection (MOD) using simulation. 

Beyond these established fields, FL has shown promising potential within UAV systems~\cite{Farsi_2024}. For example, \cite{10.1145/3477090.3481051} explored its use in drone-based collaborative learning. By employing a set of drones that communicate over a wireless network, the study successfully trained a U-Net model for collision avoidance and landing assistance, showcasing  potential of FL in edge-based, real-time applications.

Decentralized FL approaches such as {Decentralized Cyclic Weight Transfer} (\texttt{DCWT}) and {Swarm Learning} (\texttt{SL}) eliminate the dependency on a central server and shift the administration tasks to one of the clients. \texttt{DCWT} follows a sequential training process where model weights are transferred from one client to the next in a predefined order. Each client trains the model locally before passing the updated weights to the next client, forming a cyclic training approach. \texttt{SL} on the other hand, lets all the clients train simultaneously with one client being responsible for the administration tasks~\cite{Warnat-Herresthal2021}.  

Recent attention has also been drawn to the environmental implications of FL. Studies indicate that FL can be more energy-intensive and generate a larger carbon footprint compared to traditional Centralized Learning (\texttt{CL}) methods~\cite{10.5555/3648699.3648828}. The carbon emissions in FL is not only related to the training of the model on client devices, but also includes the energy used during client-server communication and the conversion of energy to carbon emissions depending on regional grid characteristics ~\cite{savazzi2022energy}. Qiu, X. et al.~\cite{10.5555/3648699.3648828} showed  that communication costs between clients and servers can represent anywhere from 0.7\% to more than 96\% of total emissions in FL systems. Farsi, A.A. et al.~\cite{fengassessing} proposed adding sustainability as a fourth pillar to the existing framework, alongside legality, ethics, and robustness. This addition addresses the environmental impacts of FL systems. Despite these concerns, direct comparative analyses focusing on energy efficiency across various FL algorithms remain limited.

This work builds on these foundations by evaluating FL in a real-world scenario of UAV-based thermal anomaly detection. It examines the performance of centralized and decentralized FL workflows and compares their effectiveness in terms of metrics such as model accuracy, energy consumption, and communication overhead. By using real-world data from two cities with significant heterogeneity, this study aims to bridge the gap between theoretical FL advances and practical applications.

\section{Material and Methods}\label{sec2}
\subsection{Dataset}

A multispectral image dataset from~\cite{10858702} forms the basis of this study, publicly available on Zenodo~\cite{vollmer_2025_10814413}. It consists of \textbf{793 images} acquired through 14 UAV flights in Germany, the large majority of which stem from Munich, the rest from Karlsruhe. All flights were carried out in nadir (\SI{90}{\degree} pitch angle) and at \SI{60}{\metre} height.

\begin{figure}[!t]
     \centering
    \begin{subfigure}[b]{0.24\textwidth}
         \centering
         \includegraphics[width=\textwidth]{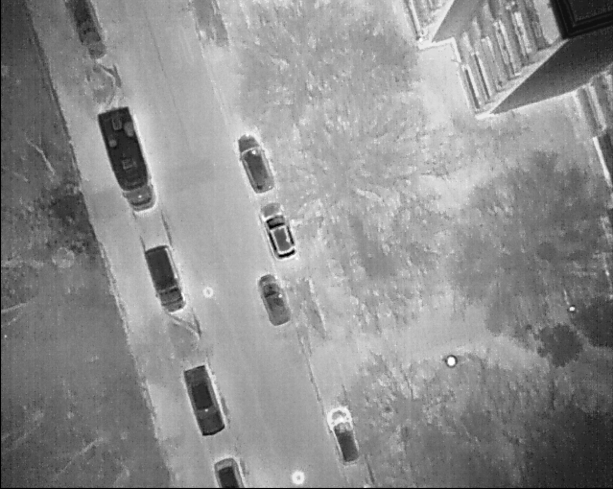}
         \caption{TIR image}
         \label{fig:ex/sub:tir}
     \end{subfigure}
     \hfill
     \begin{subfigure}[b]{0.24\textwidth}
         \centering
         \includegraphics[width=\textwidth]{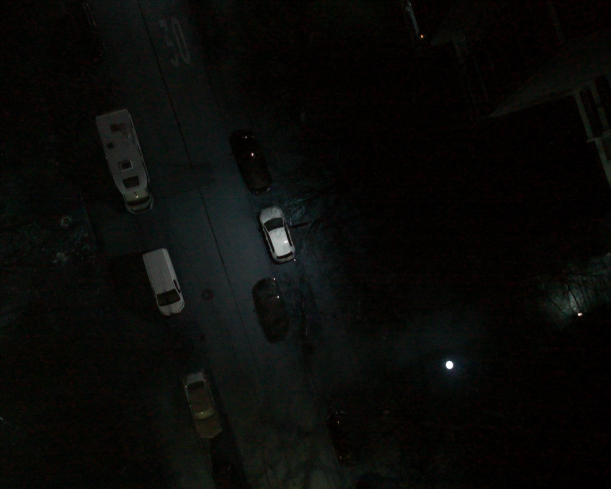}
         \caption{RGB image}
         \label{fig:ex/sub:rgb}
     \end{subfigure}
     \hfill
     \begin{subfigure}[b]{0.38\textwidth}
         \centering
         \includegraphics[width=\textwidth]{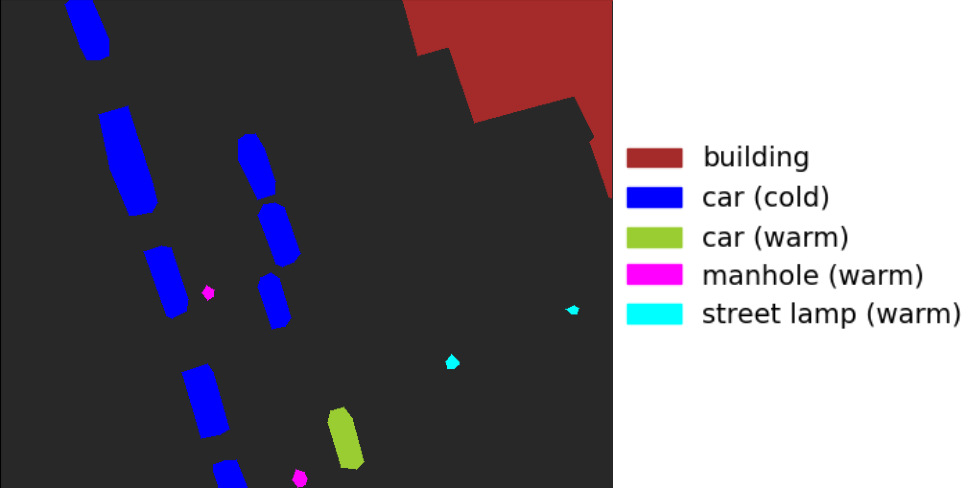}
         \caption{Annotation mask}
         \label{fig:ex/sub:anno}
     \end{subfigure}
     \caption[]{Multispectral (RGB/TIR) image and annotations example~\cite{10858702}.}
    %\caption[]{Example of a multispectral image from the dataset, consisting of RGB and TIR, as well as associated annotation mask. Figure based on~\cite{10858702}.}
     \label{fig:example-image}
\end{figure}

Both thermal infrared (TIR) and standard red-green-blue (RGB) imagery was simultaneously captured using DJI’s Zenmuse XT2, a combined 4k RGB camera and FLIR thermal sensor~\cite{10858702}. Registration procedures help compensate for differing aspect ratios, resolutions, and fields of view to create the multispectral dataset~\cite{10858702}. As the focus lies on hot-spot detection, the TIRs were annotated with \textbf{seven classes} of common thermal urban features. The classes are summed up in Table~\ref{tab:object_counts}. An example multispectral image and associated annotation are shown in Figure~\ref{fig:example-image}. Both clearly highlight the strong imbalance of instances and pixel amounts per class~\cite{10858702}.

\begin{table}[ht]
    \centering
    \caption{Aggregated object counts for KA and MU datasets~\cite{10858702}}
    \begin{tabular}{lccc}
    \toprule
    \textbf{General} & \textbf{Client-1 (MU)} & \textbf{Client-2 (KA)} & \textbf{Total (KA+MU)} \\
    \toprule
    %No. UAV flights & 12 & 2 & 14 \\
    No. images & 700 & 93 & 793 \\
    \midrule
  %  \textbf{Object Class} & & & %\\
   % \midrule
    Building & 1215 & 189 & 1404 \\
    Car (cold) & 838 & 1694 & 2532 \\
    Car (warm) & 86 & 950 & 1036 \\
    Manhole (cold) & 25 & 495 & 520 \\
    Manhole (warm) & 82 & 1297 & 1379 \\
    Miscellaneous & 81 & - & 81 \\
    Person & 275 & - & 275 \\
    Street lamp (cold) & 5 & 95 & 100 \\
    Street lamp (warm) & 11 & 672 & 683 \\
    \bottomrule
    \end{tabular}
    \label{tab:object_counts}
\end{table}

\subsection{Thermal Urban Feature Semantic Segmentation}
During the detection of thermal anomalies in UAV images, the majority of false positives arise from the common urban features listed in Table~\ref{tab:object_counts}. The U-Net model is employed to identify these features and eliminate them, thus avoiding their misclassification as potential district heating network leaks~\cite{10858702}. U-Net is a convolutional neural network architecture designed especially for image segmentation tasks~\cite{10.1007/978-3-319-24574-4_28}. 
Its distinctive U-shaped structure consists of the main parts of an encoder and a decoder, connected by skip connections.
The encoder, or contracting path, follows the typical architecture of a convolutional network, while the decoder recovers spatial information and generates the predicted segmentation mask.
In this work, a ResNet-152~\cite{He2015} encoder was used, also referred to as the backbone of the U-Net.

\subsection{Framework}
In order to apply the FL methods and algorithms and to transform an existing centralized ML workflow into a FL one, we use NVIDIA FL Application Runtime Environment
(NVFlare)~\cite{NVFlare} in this work. 

NVFlare is an open source framework for implementing FL workflows in Python. Apart from enabling the transformation of existing ML/DL workflows into a FL paradigm, it offers a wide range of features, which include: simulation and prototyping tools, privacy-preserving algorithms, built-in FL algorithms, productivity enhancement (like MLFlow~\cite{mlflow}), provisioning tools for participant verification and secure communication, and many more~\cite{NVFlare}. %These features make NVFlare a robust solution for researchers and industry applications seeking to implement FL at scale while maintaining security and privacy~\cite{Riedel2024}.

\subsection{Aggregation Algorithms and Workflows}
\label{chp3:sec4:algorithms}
The most commonly used aggregation algorithms in FL, including \texttt{FedAvg}, \texttt{FedProx}, \texttt{FedOpt}, and \texttt{Scaffold}, were selected for this work. \texttt{FedAvg} is the first and simplest algorithm introduced in FL, with others developed to address its drawbacks regarding the non-IID nature of datasets in FL settings. %At the time of these experiments, these algorithms were available in the NVFlare framework for PyTorch. However, except for \texttt{FedAvg}, none were provided for the TensorFlow framework. We implemented these algorithms from scratch and contributed our implementations to the framework~\cite{pullRequest}.

As mentioned before, \texttt{FedAvg} uses weighted averages when aggregating local weights sent back from the clients.  

\texttt{FedProx} introduces a proximal term \( \frac{\mu}{2} \|w - w^t\|^2 \) to the local objective function, where \( w \) is the local model  weights, \( w^t \) is the global model weights, and \( \mu \) controls regularization. 

\texttt{FedOpt} uses stochastic gradient descent (SGD) to update the global model. The global gradient update is:

\[
\Delta_t = \frac{1}{|S|} \sum_{i \in S} (w_{i,K}^t - w^t)
\]

where \(S\) is the set of participating clients, \(w_{i,K}^t\) is the local model after \(K\) training steps, and \(w^t\) is the global model weights in round \(t\). The global model is then updated as:
\[
w^{t+1} = \textsc{ServerOpt}(w^t, -\Delta^t, \eta, t)
\]
where \(\eta\) is the server learning rate and \textsc{ServerOpt} refers to the optimization function or algorithm applied by the server to update the global model.

\texttt{Scaffold} is an enhanced FL algorithm that mitigates client drift caused by data heterogeneity using control variates. The global update is given by:
\begin{align}
    w^{t+1} &= w^t + \eta \frac{1}{|S|}\sum_{i \in S}(\Delta_i^t - c_i^t + c^t),\quad c^t=\frac{1}{N}\sum_{i=1}^{N} c_i^t
\end{align}
where \(w^t\) is the global model in round $t$, \(\Delta_i^t\) the local update, \(c_i^t\) and \(c^t\) are client and global control variates, \(\eta\) the learning rate, and \(|S|\) the number of participating clients.

When FL was first introduced, it initially consisted of a Scatter \& Gather (S\&G) workflow, relying on a central server for aggregation.%It includes several clients for training and a server for administrative tasks. In a scatter step, the global model weights are sent to training clients. This shared model is trained locally by the clients and the updated model weights are in turn collected by the server as part of the gathering part of the workflow.
%When the server has received all the updated weights, it aggregates these updates based on one aggregation algorithm and creates a new global model. This final step completes one round of training in a FL workflow. The aggregated model weights are sent to the client for the next round of training. This process continues until the predefined number of rounds is reached.

When there is no trusted server available, Decentralized Federated Learning (DFL) such as Decentralized \texttt{CWT} (\texttt{DCWT}) and Swarm Learning (\texttt{SL}) can be used as alternatives. In such cases, communication is conducted peer-to-peer. Since there is no server to be compromised, DFL provides enhanced privacy protection compared to centralized approaches and it reduces network communication. In \texttt{SL}, one of the clients takes responsibility for performing aggregation. In the implementation of these workflows in NVFlare, a server is still present to handle administrative tasks, but no sensitive data is shared with it.

\subsection{Training Setup}

Training the semantic segmentation model requires considerable computational resources, so each client was assigned a GPU node on a HPC system. \textbf{Client-1} utilized a GPU node on HoreKa~\cite{HoreKA}, while \textbf{Client-2} used a GPU node on HAICORE~\cite{HAICORE}. Both nodes provide the same resources, including \texttt {2 Intel Xeon Platinum 8368} processors with a combined \texttt {76 CPU Cores} and \texttt{152 CPU Threads}, \texttt{512 GB main memory}, \texttt{4 NVIDIA A100-40 GPUs} with \texttt{40 GB Memory} each and a local \texttt{NVMe SSD disk} of \texttt{960 GB size}. However, only one GPU from the four GPUs available on each node was used for each client. 
To run NVFlare clients on our HPC cluster, we used the SLURM job scheduler~\cite{slurm}. Using NVFlare's provisioning function several bash scripts for launching clients and the server are created. We modified these scripts to make them compatible with SLURM, enabling deployment on the HPC systems.

Since our dataset consists of images from two cities, our initial configuration was a simple location-based split with one client per city, as shown in Table~\ref{tab:object_counts}. To evaluate the scalability another setup including five clients was tested. For this the data was splitted into five heterogeneous datasets to measure the performance of the FL algorithms with an increased number of clients while maintaining the same overall dataset size. In this setup, due to resource constraints, four clients were launched as a SLURM job, each running on the same node with a dedicated GPU. The fifth client was started with a separate SLURM job to prevent GPU memory allocation issues during training. Table~\ref{tab:dataset_counts} shows the object count and number of images per client. Still, the class count varies across the clients, and this significant variance indicates that the underlying distributions are not identical. %The KA dataset is a subset of the dataset used by Client-2. 

\begin{table}[ht]
\centering
\caption{Object counts for 5-clients}
\begin{tabular}{lccccc}
\toprule
\textbf{General} & \textbf{Client-1} & \textbf{Client-2} & \textbf{Client-3} & \textbf{Client-4} & \textbf{Client-5} \\
\toprule
No. images & 159 & 159 & 160 & 164 & 151 \\
\midrule
%\textbf{Object class} & & & & & %\\
%\midrule
Building & 181 & 459 & 294 & 213 & 257 \\
Car (cold) & 231 & 1093 & 303 & 165 & 740 \\
Car (warm) & 72 & 123 & 181 & 302 & 358 \\
Manhole (cold) & 135 & 74 & 80 & 144 & 87 \\
Manhole (warm) & 367 & 166 & 293 & 287 & 266 \\
Miscellaneous & 135 & 74 & 80 & 144 & 87 \\
Person & 0 & 0 & 85 & 137 & 53 \\
Street lamp (cold) & 45 & 6 & 19 & 13 & 17 \\
Street lamp (warm) & 118 & 131 & 159 & 145 & 130 \\
\bottomrule
\end{tabular}
\label{tab:dataset_counts}
\end{table}

The server was set up on the bwCloud provided by KIT. The server ran as a Virtual Machine (VM) with \texttt{32 GB main memory} and \texttt{40 GB storage}~\cite{bwcloud}.

%MLflow ~\cite{mlflowdocs} is an open-source platform designed to manage the lifecycle of ML/DL models. It provides tools for tracking experiments, packaging code into reproducible runs, and sharing and deploying models across diverse environments.
Before sending a model to production, it is important to track and monitor it in order to optimize and check its quality. To this end, MLOps is an engineering practice that aims to automate and streamline the ML lifecycle. For this purpose, in this work MLflow was chosen for experiment tracking as it is supported by NVFlare, and an MLflow tracking server in production was provided by the AI4EOSC project~\cite{ai4eosc}. During training, each client sends local metrics and parameters to the NVFlare server, which logs them on the configured MLflow server. By default, NVFlare does not log the global model in MLflow, so we modified NVFlare source code to enable this functionality for each experiment.

The  metrics chosen for evaluating the model consist of the {mean accuracy} ($\mathrm{mACC}$), {mean weighted precision ($\mathrm{mwP}$)}, {mean weighted $F_1$-score ($\mathrm{mwF_1}$)} and {mean weighted Intersection Over Union score ($\mathrm{mwIoU}$)}. The weights in these metrics are determined by the number of pixels occupied by each specific class within the annotation masks.

%TODO: Add a comment about scalability and connect to the simulation with five clients

The thermal urban feature semantic segmentation use case, originally developed and trained in a centralized way with predefined hyperparameters, was adapted into a FL workflow. The training hyperparameters identified by~\cite{Vollmer2024_AI} were already optimized and considered ideal for this application. The parameters for the FL workflow were aligned with those of the \texttt{CL} workflow. The sigmoid focal cross-entropy loss function, Adam optimizer,  a learning rate (lr) of $0.001$ were used~\cite{Vollmer2024_AI,10858702}. Experiments used $2$ or $5$ clients, with $4$ rounds for two clients and $4$, $7$, or $13$ for five clients. Local epochs were $18$ for \texttt{CL}, $5$ or $9$ for five clients, and batch size was fixed at $8$. As is~\cite{10858702}, the dataset was split into 80\% training and 20\% testing. As shown in Table~\ref{tab:dataset_counts}, the dataset is imbalanced which is addressed by the authors in greater detail in \cite{10858702}—who apply techniques such as transfer learning and tailored loss functions and class-balanced metrics (e.g., weighted precision, $F_1$, IoU). Since we used the same model and configuration as in \cite{10858702}, the data imbalance is already addressed.

%TODO Mention again that the same configuration is used as in the orginal work(model, hyperparameters, metrics etc.)

To efficiently calculate the energy consumption of each algorithm, we utilized the \textit{perun} package to track the power draw of CPU, GPU, and memory of both clients and server~\cite{10.1007/978-3-031-39698-42}. Since our original server was set up on a VM, we could not efficiently track the energy consumption due to the lack of direct access to hardware performance counters. To overcome this limitation, we set up the server on bare-metal equipped with hardware energy measurement capabilities (such as Intel RAPL or a similar power monitoring framework) on a dedicated laptop. This setup allowed us to obtain more accurate energy consumption data for our algorithms and compare them.

\section{Results and Discussions}\label{sec3}
The model is trained using the available images to automatically detect and eliminate false alarms.
The script for this workflow is written in Python 3.8 and uses Tensorflow 2.10. The results of training multiple models using the various approaches introduced in Section~\ref{chp3:sec4:algorithms} are presented and discussed in this section. 
 
\subsection{Comparison of the Aggregation Algorithms}
This section presents the results of applying different FL aggregation algorithms, specifically \texttt{FedAvg}, \texttt{FedProx}, \texttt{FedOpt}, and \texttt{Scaffold}. Each experiment was conducted five times to ensure statistical significance. We first trained the model with two clients (KA and MU dataset). This experiment is referred to as the \emph{2-client scenario}. Then, the dataset was divided into five equivalent subsets, and the experiment was repeated to investigate the results when all clients had datasets of equal size within the \emph{5-client scenario}.

In \texttt{FedOpt} and \texttt{Scaffold}, where server-side optimization updates trainable weights, performance dropped after two rounds and continued to decline with each subsequent round when using standard Batch Normalization (BN) layers \cite{NEURIPS2018_36072923} in the U-Net backbone, despite weighted averaging of non-trainable parameters like running mean and variance. In BN, locally updated running statistics vary across clients, leading to inconsistencies that degrade global model performance and stability when aggregated. To address this, we replaced BN with Group Normalization (GN)~\cite{wu2018groupnormalization}, which normalizes within channel groups instead of batches, making it independent of batch size and more robust for FL. Moreover, since \texttt{FedOpt} uses SGD to update the global model, \texttt{lr} and \texttt{momentum} can be tuned to optimize the update. NVFlare uses default \texttt{lr} $1.0$ and \texttt{momentum} of $0.9$, and our experiments showed that reducing \texttt{lr} below $1.0$ leads to a performance drop in the global model performance. We therefore adhered to the default values. 

Table~\ref{tab:comparisonSG-2clients} shows the results of training the model within the \emph{2-client scenario}. The first number reported for each client represents the average metric value over five executions of the algorithm, while the second number denotes the standard uncertainty. For the overall results, we computed a weighted average of each metric from both Client-1 and Client-2. 

The global model aggregated using \texttt{FedAvg} achieved the highest performance across all metrics on the local test set of Client-1 (MU). It recorded a mean weighted precision of $0.945\pm0.003$, a mean accuracy of $0.94\pm 0.007$, and an $F_1$ score of $0.937\pm 0.005$. On the other hand, \texttt{FedProx} showed the weakest performance for this client, particularly in terms of mean accuracy $0.92\pm0.01$ and mean weighted IoU ($0.85\pm 0.02$ ). When comparing \texttt{FedAvg} with \texttt{Scaffold}, \texttt{FedAvg} shows slightly better performance in terms of $\mathrm{mACC}$,  $\mathrm{mwF_1}$, and  $\mathrm{moU}$. \texttt{FedOpt} consistently shows the weakest performance for Client-1, with values that often fall below other ranges of algorithms.

The global model aggregated using \texttt{Scaffold} demonstrated the best performance among FL algorithms on Client-2 and has the closest performance to \texttt{CL}.  It surpasses \texttt{FedAvg} in both $\mathrm{mACC}$ ($0.846 \pm 0.009$ vs. $0.66 \pm 0.007$) and $\mathrm{mwF_1}$ ($0.858 \pm 0.007$ vs. $0.810 \pm 0.014$). Additionally, \texttt{Scaffold} $\mathrm{mwIoU}$ of $0.777 \pm 0.009$ shows an 
improvement over \texttt{FedAvg} $0.737 \pm 0.014$.  \texttt{FedProx} consistently shows the weakest performance for Client-2, with the widest error ranges indicating less stability.

When comparing \texttt{CL} with FL approaches based on overall results, the model trained using the \texttt{CL} method and the global model aggregated using \texttt{FedAvg} and \texttt{Scaffold} show comparable performance across several key metrics. For instance, \texttt{FedAvg} achieves a $\mathrm{mwP}$ of $0.935 \pm 0.005$, which is statistically consistent with \texttt{CL} with  $\mathrm{mwP}$ of  $0.939 \pm 0.002$. These results highlight that \texttt{FedAvg} and \texttt{Scaffold} can achieve performance comparable to centralized training while preserving data privacy. When considering error ranges, \texttt{Scaffold} and \texttt{FedAvg} can match or even exceed \texttt{CL} in specific metrics. Overall, this highlights their potential as effective FL strategies for our application. In contrast, \texttt{FedProx} and \texttt{FedOpt} show lower performance across several metrics, suggesting that while they can be effective, they may be more sensitive to client data variability.

\begin{table}
\centering
\small
\renewcommand{\arraystretch}{0.9} 
\caption{Performance of the global model trained with different FL Algorithms and \texttt{CL} on the test set of Client-1 (MU), Client-2 (KA), and (MU+KA)}\label{tab:comparisonSG-2clients}
\begin{tabular}{clccccc}

\toprule
&\textbf{Metric} &  \textbf{Centralized}& \textbf{FedAvg} & \textbf{FedProx} & \textbf{FedOpt} & \textbf{Scaffold} \\
\midrule
\multirow{4}{*}{\rotatebox[origin=c]{90}{{\textbf{Client-1}}}}
&\rowheight\text{$\mathrm{mwP}$} & $\bm{0.952\pm0.017}$  & $ 0.945 \pm 0.003$  & $ 0.924 \pm 0.007$  & $0.913 \pm 0.001$ & $0.938 \pm 0.001$\\  
&\rowheight\text{$\mathrm{mACC}$}          & $\bm{0.949\pm 0.003}$ &$0.940 \pm 0.007$ &$0.911 \pm 0.006$&  $ 0.879 \pm 0.009$ & $0.923 \pm 0.004$  \\  
&\rowheight $\mathrm{mwF_1 }$           & $\bm{0.944\pm 0.002}$ & $0.937 \pm 0.005$& $ 0.903 \pm 0.009$ &$ 0.887 \pm 0.005$ &  $ 0.924 \pm 0.002$\\  
&\rowheight $\mathrm{mwIoU }$      &$\bm{0.907\pm 0.005}$ &$0.900 \pm 0.007$ &$0.850 \pm 0.01$ &$0.822 \pm 0.009$ & $ 0.877 \pm 0.004$\\  
\midrule

\multirow{4}{*}{\rotatebox[origin=c]{90}{\textbf{Client-2}}}
&\rowheight\text{$\mathrm{mwP}$} &  $0.870\pm0.015$ & $ 0.853 \pm 0.012$& $ 0.834 \pm 0.025$ &$0.843 \pm 0.005$
& $\bm{0.879 \pm 0.003}$ \\  
&\rowheight\text{$\mathrm{mACC}$} & $0.866\pm 0.012$ & $0.846 \pm 0.009$&$ 0.818 \pm 0.017$& $0.833 \pm 0.005$ &$\bm{0.866 \pm 0.007}$ \\  
&\rowheight $\mathrm{mwF_1 }$       &$0.847\pm 0.017$& $0.810 \pm 0.014$ & $0.779 \pm 0.031$  & $ 0.818 \pm 0.004$& $\bm{ 0.858\pm 0.007}$\\  
&\rowheight $\mathrm{mwIoU }$      &  $0.774\pm 0.018$ & $0.737\pm 0.014$& $0.702 \pm 0.029$  &$0.733 \pm 0.005$ & $\bm{0.777 \pm 0.009}$\\  
\midrule

\multirow{4}{*}{\rotatebox[origin=c]{90}{\textbf{Overall}}}
&\rowheight\text{$\mathrm{mwP}$} &  $\bm{0.939 \pm 0.002}$ & $0.935 \pm 0.005$  &$0.913 \pm 0.011$& $0.905 \pm 0.002$ &$0.931 \pm 0.001$ \\  
&\rowheight\text{$\mathrm{mACC}$}           & $\bm{0.931 \pm 0.002}$ & $ 0.930 \pm 0.007$&$ 0.900 \pm 0.008$ &$0.873 \pm 0.008$  &  $ 0.917 \pm 0.004$ \\  
&\rowheight $\mathrm{mwF_1 }$      & $\bm{ 0.928 \pm 0.002}$ & $0.922 \pm 0.007$ &  $0.888 \pm 0.013$ & $0.879 \pm 0.006$ &  $0.916 \pm 0.003$ \\  
&\rowheight  $\mathrm{ mwIoU}$      & $\bm{0.884 \pm 0.003}$ & $ 0.881 \pm 0.008$ & $0.833 \pm 0.014$ & $0.812 \pm 0.009$ &  $0.865 \pm 0.005$\\   
\bottomrule
\end{tabular}

\end{table}

In the subsequent experiments, we used the perun package~\cite{10.1007/978-3-031-39698-42} to measure the energy consumption of each FL algorithm during the training process. It also offers a calculation of an average carbon intensity of electricity based on a emissions factor which can differ per region and can be set manually by the user. This calculation was not reported within this work due to being highly versatile between different regions of the world.
%TODO: default average value or based on location?
To ensure the reliability and consistency of our results, we took specific precautions to
eliminate potential interference. We requested exclusive access to the HPC nodes where our clients were executed, ensuring that no other jobs were running on the
same nodes during the experiments. This isolation minimized resource contention and provided a controlled environment for energy measurement. In the \emph{5-client scenario}, four clients shared the same node, making it challenging to measure the energy consumption of each client individually. Therefore, we conducted energy
consumption measurements for only two clients. For \texttt{CL}, we used the same dedicated node on HoreKa as for the FL clients, utilizing one of the four available GPUs.

\begin{figure}[ht]  
    \centering  
    \includegraphics[width=0.75\linewidth]{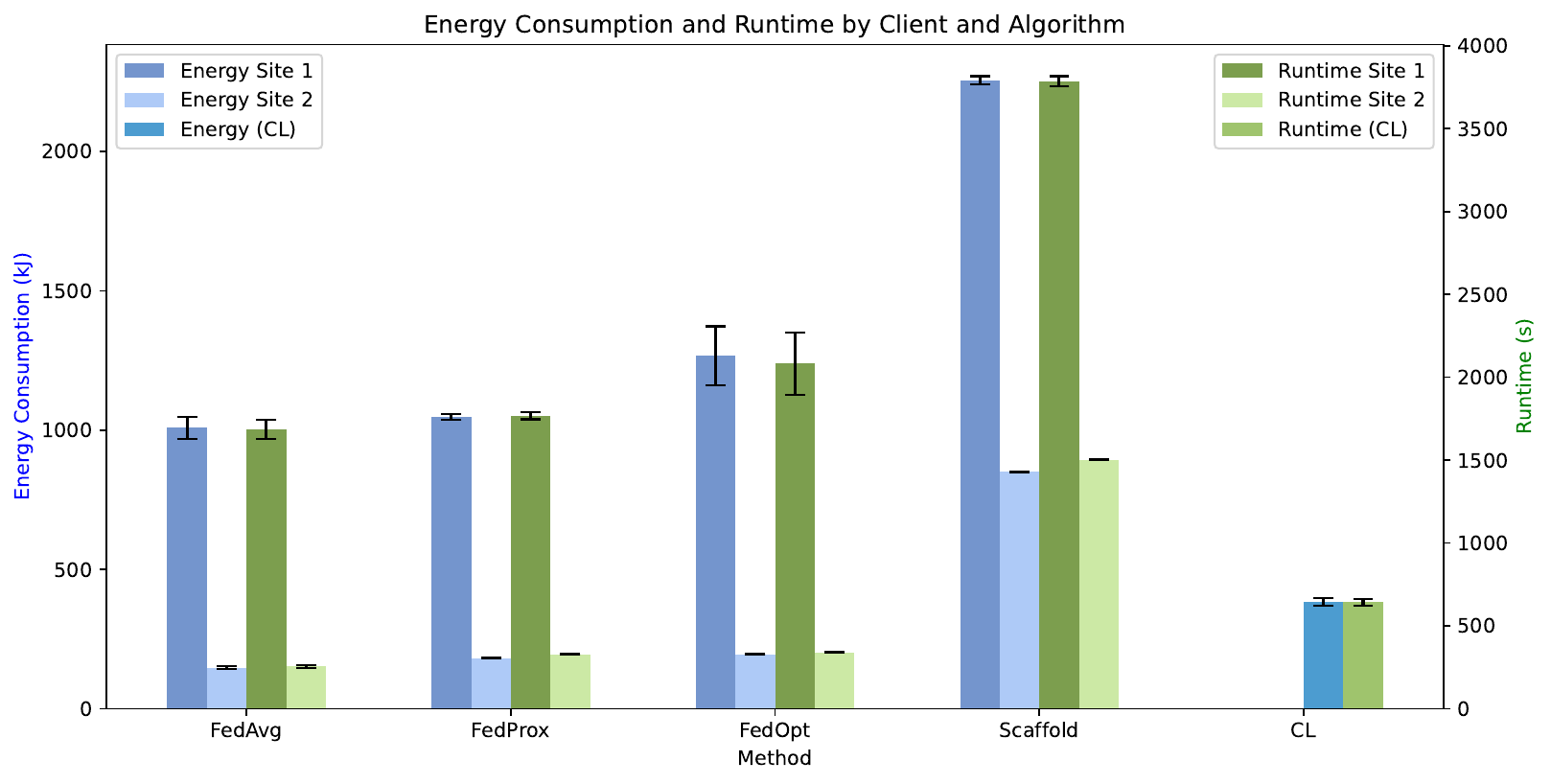}  
    \caption{Energy consumption and runtime by each site and each algorithm.}
    \label{fig:example_figure}  
\end{figure}

The energy consumption on the server for each algorithm was very low (14–29 \si{\kJ}). Figure~\ref{fig:example_figure} represents the mean
execution time and energy consumption of various FL algorithms for two clients. As shown in the figure, \texttt{Scaffold} demonstrates the highest execution time, resulting in the largest energy consumption across all sites. On average, it requires 103.33 minutes with a standard deviation of
8.84 minutes, highlighting its computational intensity. 
In our application, the energy consumption and runtime of \texttt{CL} compared to \texttt{FedAvg} show significant reductions. For energy consumption, \texttt{CL} consumes 382.279 \si{\kJ}, whereas \texttt{FedAvg} consumes 1008.9152 \si{\kJ}, resulting in a percentage reduction of approximately 163.97\%. Similarly, for runtime, \texttt{CL} takes 643.245 seconds compared to 1687.0624 seconds for \texttt{FedAvg} leading to a percentage reduction of approximately 162.3\%.

Reducing the available data per client dividing it by five leads to slower convergence. With five clients instead of two, each client contributes smaller updates per round, requiring more communication to reach the same performance. Since each client only accesses \( \frac{1}{5} \) of the full data, we ensure a fair comparison with \texttt{CL} by maintaining the ratio
$
\frac{\texttt{num\_rounds} \times \texttt{num\_local\_epochs}}{\texttt{num\_clients}} = \texttt{num\_epochs\_CL}
$.
For these reasons, we first reduced the number of local epochs to five and increased the number of rounds to 13. In another experiment, we kept the local epochs at nine and increased the rounds to seven. Both setups achieved similar performance, but with fewer epochs and more rounds, communication increased, leading to longer training time. Therefore, we settled on nine local epochs and seven rounds as a balanced choice. Table~\ref{tab:comparisonSG-5clients} shows the performance of the trained global model on the test dataset (KA+MU). When the dataset size is equal among clients, \texttt{FedProx} performed similar to \texttt{CL}. In \texttt{FedProx}, the aggregation on the server is done similarly to \texttt{FedAvg}. In the case of two clients, the global weights lean toward the MU dataset because of its larger weight. This means that when we compute the proximal term in \texttt{FedProx} the influence of the MU dataset is disproportionately high, potentially causing the corresponding client to pull the global model further away from the parameters of KA dataset. In contrast, in the \emph{5-client scenario}, the MU dataset is divided among several clients. Consequently, the global weights become more reflective of the average of all local weights. This balanced aggregation results in a smaller difference between local and global models, thereby reducing the magnitude of the proximal term. With a lower proximal penalty, local updates are less constrained, which can lead to more effective convergence and improved overall performance in \texttt{FedProx} compared to the \emph{2-client scenario}.

In this case, the performance of \texttt{Scaffold} dropped by 3-5\% on each metric compared to \texttt{CL}. This decline may be due to the reduced dataset size available to each client, which adversely affects the estimation of the control variate. 

\texttt{FedAvg}, achieving a mean weighted precision of $0.919\pm 0.012$, while \texttt{FedOpt} still performed the worst among other FL algorithms. In \texttt{FedAvg}, the MU dataset features richer annotations and includes every class. In the \emph{2-client scenario}, we assign greater weight to the MU dataset during aggregation, while the KA dataset receives less weight. However, in a \emph{5-client scenario}, the MU dataset is split across several clients, resulting in it having the same weight as the KA dataset during aggregation. This difference in weighting may be a key factor contributing to the observed performance drop with five clients compared to two.

\begin{table}[ht]
\centering
\caption{Comparison of different FL algorithms and CL (overall) for five clients.} 
\label{tab:comparisonSG-5clients}
\begin{tabular}{lccccc}
\toprule
%\multicolumn{6}{c}{\textbf{Scatter \& Gather}} \\
%\midrule
%\multirow{2}{*}{\textbf{Metric}} %& \multirow{2}{*}%{\textbf{Centralized} }& %\multicolumn{4}{c}%{\textbf{Federated Aggregation %Algorithms}} \\
%\cline{3-6}
\textbf{Metric}&\textbf{Centralized}   & \textbf{FedAvg} & \textbf{FedProx} & \textbf{FedOpt} & \textbf{Scaffold} \\
\midrule
$\mathrm{mwP}$ & $\bm{0.939 \pm 0.002}$ & $0.919 \pm 0.012$ & $0.932 \pm 0.003$ & $0.891 \pm 0.006$ & $0.900 \pm 0.002$ \\
$\mathrm{mACC}$ & $\bm{0.931 \pm 0.002}$ & $0.916 \pm 0.013$ & $0.926 \pm 0.006$ & $0.877 \pm 0.006$ & $0.879 \pm 0.004$ \\
$\mathrm{mwF_1}$ & $\bm{0.928 \pm 0.002}$ & $0.904 \pm 0.016$ & $0.918 \pm 0.008$ & $0.865 \pm 0.008$ & $0.874 \pm 0.002$ \\
$\mathrm{mwIoU}$ & $\bm{0.884 \pm 0.003}$ & $0.855 \pm 0.019$ & $0.872 \pm 0.009$ & $0.801 \pm 0.009$ & $0.810 \pm 0.003$ \\
\bottomrule
\end{tabular}
\end{table}
\subsection{Comparison of the Workflow}
In NVFlare, \texttt{CWT} workflow allows for configuring the order of clients during training. Specifically, the \texttt{cyclic\_order} parameter can be set to determine whether the sequence of clients is fixed or random for each round.  
To investigate how the order of clients affects the performance of the model, especially with two clients where the dataset sizes differ, we conducted several experiments using both fixed and random orders.

Table~\ref{tab:comparisonWorkflows} presents the overall results of training the model with two and five clients across different workflows. In \emph{2-client scenario}, the \texttt{SL} workflow achieved similar performance to S\&G (\texttt{FedAvg}), as expected: The only difference between these workflows is that in S\&G, aggregation is performed on the server, whereas in \texttt{SL}, this task is done by one of the clients. In terms of execution time, the \texttt{SL} algorithm achieves a mean execution time of 949.86s (for 2 clients), whereas \texttt{FedAvg} records a significantly higher mean execution time of 2906.07s, representing an approximate 205.95\% increase in training time. This is due to the elimination of communication overhead between clients and the server, which reduces latency and speeds up the training process in the \texttt{SL} workflow.

In \texttt{CWT}, we first train the model on the MU dataset as Client-1 and then transfer the  weights of the model to Client-2 (KA dataset). In the next experiment, we also perform training in the reverse order, first on Client-1 (KA dataset) and then on Client-2 (MU dataset). The results of these experiments are presented in Table~\ref{tab:comparisonWorkflows}: We observe that training the model first on the KA dataset followed by MU (\texttt{CWT KA-MU}) consistently yields better performance across all metrics compared to training in the reverse order (\texttt{CWT MU-KA}). This suggests that KA with less data  provides a stronger initial model foundation, leading to better generalization when fine-tuned on MU. Conversely, training on the larger MU dataset first, then on KA, can cause the model to overfit the smaller KA dataset.

In the case of five clients, we used a random order, and the performance was similar to that of the model trained with a fixed order (Client-1 to Client-5). Therefore, in \texttt{CWT}, the order of the clients should be carefully considered when the dataset sizes differ, as it can significantly impact the final model performance.

Furthermore, comparing different workflows, \texttt{SL} outperforming all the \texttt{CWT} and \texttt{DCWT} variations. This indicates that local aggregation in \texttt{SL} may provide better feature learning compared to \texttt{CWT}.

As shown in Table~\ref{tab:comparisonWorkflows}, for the \emph{5-client scenario}, all frameworks exhibit statistically consistent performance across key metrics. Considering error ranges, the results indicate no significant differences among the methods, suggesting that each approach is capable of delivering comparable results despite variations in their training and aggregation processes.

\begin{figure}[ht]  
    \centering  
    \includegraphics[width=0.75\linewidth]{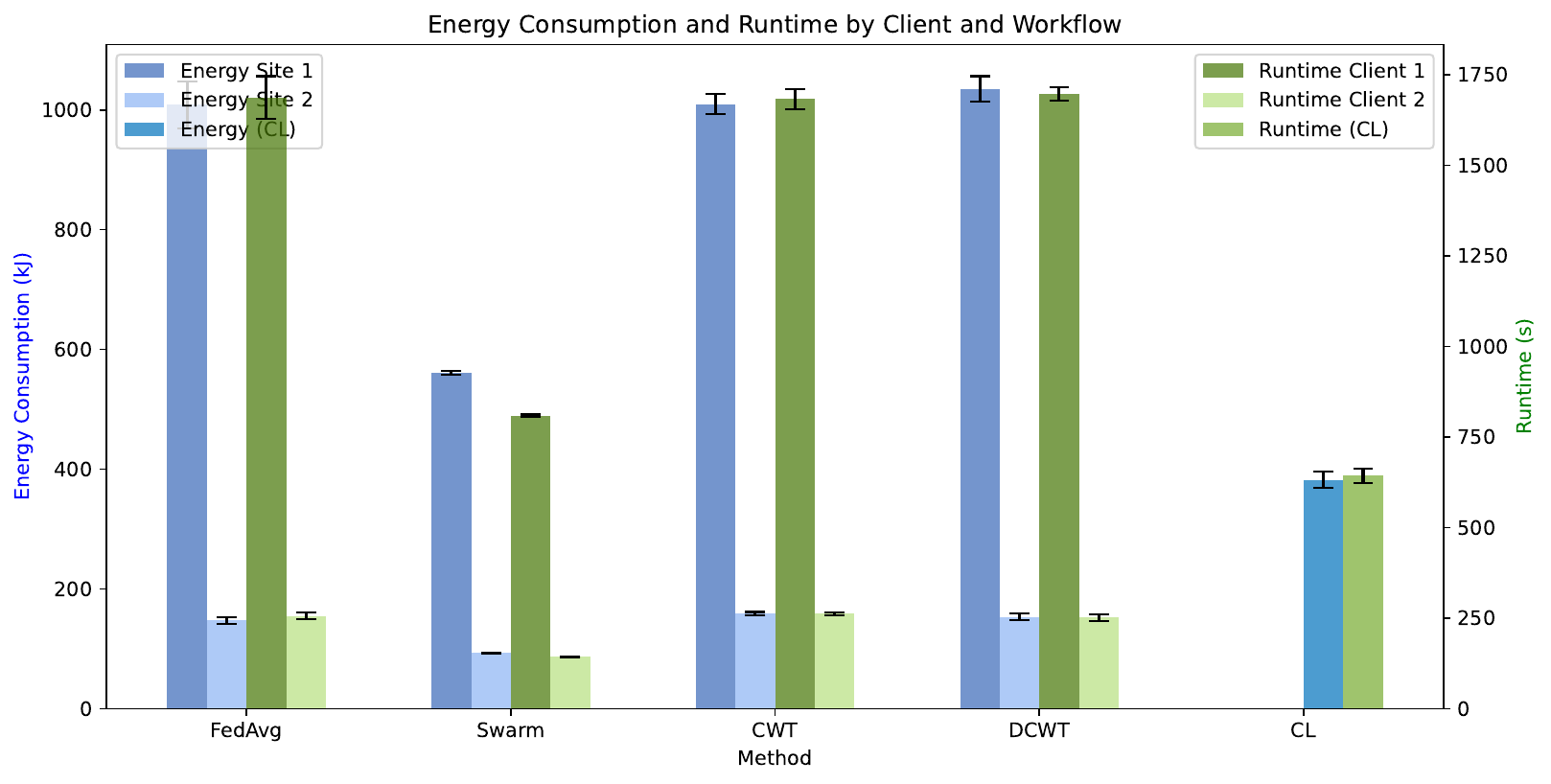}  
    \caption{Energy consumption and runtime by each site and each workflow.}
    \label{fig:example_figure1}  
\end{figure}

\begin{table}[!t]
 \renewcommand{\arraystretch}{0.85}
\centering
\caption{Comparison of Different Workflows in FL.}
\label{tab:comparisonWorkflows}
\resizebox{\columnwidth}{!}{%
\begin{tabular}{@{}lcccccc|ccc@{}}
\toprule
 & \multicolumn{6}{c|}{\textbf{Two Clients}} & \multicolumn{3}{c}{\textbf{Five Clients}} \\
\midrule
\textbf{Metric} & \textbf{S\&G} & \textbf{Swarm} & \textbf{CWT} & \textbf{CWT} & \textbf{DCWT} & \textbf{DCWT} & \textbf{S\&G} & \textbf{Swarm} & \textbf{CWT} \\
 & (FedAvg) & Learning & (MU-KA) & (KA-MU) & (MU-KA) & (KA-MU) & (FedAvg) & Learning & Random \\
\midrule
$\mathrm{mwP}$ & $0.935$ & $\mathbf{0.940}$ & $0.924$ & $0.933$ & $0.917$ & $0.932$ & $0.919$ & $\mathbf{0.939}$ & $0.935$ \\
 & $\pm 0.023$ & $\mathbf{\pm 0.020}$ & $\pm 0.035$ & $\pm 0.009$ & $\pm 0.026$ & $\pm 0.030$ & $\pm 0.012$ & $\mathbf{\pm 0.003}$ & $\pm 0.005$ \\
\midrule
$\mathrm{mACC}$ & $0.930$ & $\mathbf{0.940}$ & $0.858$ & $0.895$ & $0.866$ & $0.913$ & $0.916$ & $\mathbf{0.934}$ & $0.926$ \\
 & $\pm 0.033$ & $\mathbf{\pm 0.021}$ & $\pm 0.127$ & $\pm 0.095$ & $\pm 0.056$ & $\pm 0.015$ & $\pm 0.013$ & $\mathbf{\pm 0.003}$ & $\pm 0.004$ \\
\midrule
$\mathrm{mF_1}$ & $0.922$ & $\mathbf{0.930}$ & $0.875$ & $0.916$ & $0.874$ & $0.912$ & $0.904$ & $\mathbf{0.923}$ & $0.919$ \\
 & $\pm 0.030$ & $\mathbf{\pm 0.027}$ & $\pm 0.096$ & $\pm 0.012$ & $\pm 0.043$ & $\pm 0.046$ & $\pm 0.016$ & $\mathbf{\pm 0.005}$ & $\pm 0.005$ \\
\midrule
$\mathrm{mwIoU}$ & $0.881$ & $\mathbf{0.892}$ & $0.796$ & $0.864$ & $0.802$ & $0.860$ & $0.855$ & $\mathbf{0.883}$ & $0.875$ \\
 & $\pm 0.037$ & $\mathbf{\pm 0.034}$ & $\pm 0.148$ & $\pm 0.021$ & $\pm 0.060$ & $\pm 0.056$ & $\pm 0.019$ & $\mathbf{\pm 0.005}$ & $\pm 0.006$ \\
\bottomrule
\end{tabular}%
}
\end{table}
Figure~\ref{fig:example_figure1} represents the mean
execution time and energy consumption of various FL workflows for two sites. \texttt{SL} records a mean time execution of 952.42s
whereas \texttt{DCWT} records a significantly higher mean execution time of 1949.69s. This represents an approximate 105\% increase in training time. In \texttt{DCWT}, clients train the model sequentially, requiring each client to wait for the previous one to complete its training. As the number of clients increases, this sequential process leads to longer overall training times. Again, comparing \texttt{DCWT} to \texttt{CWT}, when communication time with the server is removed, \texttt{DCWT} demonstrates improved efficiency.
\section*{Conclusions}
In this work, we successfully transformed the introduced application from a traditional ML workflow into a FL workflow. The results show that FL algorithms can achieve comparable performance metrics to CL in detecting small objects within UAV images. \texttt{FedProx} shows variability in performance depending on client configurations. It performs poorly in the \emph{2-client scenario} but substantially improves and achieves comparable results to CL in the \emph{5-client scenario} due to reduced proximal penalties when data is more evenly distributed. \texttt{FedOpt} introduces additional hyperparameters, making it more challenging to set them optimally without prior data insights. \texttt{Scaffold}, on the other hand, can enhance performance but requires increased execution time and energy consumption, and in some cases, the improvements may not be significant enough to justify the additional cost.  When the number of clients increases, convergence slows due to less data per client. Adjusting the training strategy by balancing the number of local epochs and aggregation rounds helps achieve optimal trade-offs between performance and computational cost.  

Despite these promising findings, some limitations remain. Due to limited access to real-world data, scalability was tested only with a simulated setup of five clients, which may not capture real-world complexity. Moreover, while our study focuses on evaluating performance and system-level efficiency across various FL workflows, model interpretability was not within scope. In future work this aspect should be considered.
%TODO: add model interpretabilty in future work

We also observed practical system-level challenges. Although HPC systems offer additional computational resources, batch job scheduling limits client availability, causing some clients to lack resources when others are ready.

In terms of sustainability, in our case study, when we compared to centralized training, experimental evidence shows FL can consume more energy than its centralized version in certain scenarios. Introducing decentralized workflows like \texttt{SL} and \texttt{DCWT} show a gain in shorter training time. Among FL workflows, 	\texttt{SL} demonstrates superior computational efficiency and runtime performance compared to S\&G and 	\texttt{CWT} approaches. However, this relationship is not universal.It is highly depending on the setup, including the deployed machines, used algorithm/workflow and even the geographical region of the participating devices. 

\section*{Acknowledgment}
This work is supported by the AI4EOSC project, which receives funding from the European Union's Horizon Europe 2022 research and innovation programme under agreement 101058593. Additionally, computational resources were provided by the HoreKa supercomputer, funded by the Ministry of Science, Research and the Arts Baden-Württemberg and the Federal Ministry of Education and Research, and by the Helmholtz Association Initiative and Networking Fund through the Helmholtz AI platform grant and the HAICORE@KIT partition.

This preprint has not any post-submission improvements or corrections. The Version of Record of this contribution is published in Computational Science and Its Applications – ICCSA 2025, and is available online at https://doi.org/10.1007/978-3-031-97000-9\_18

\bibliography{custom}
\bibliographystyle{splncs04}
\end{document}